%% file: main.tex
\documentclass[10pt,twocolumn,letterpaper]{article}

\usepackage{cvpr}              

\usepackage{amsmath,amssymb,amsfonts}
\usepackage{algorithmic}
\usepackage{graphicx}
\usepackage{textcomp}
\usepackage{xcolor}
\usepackage{cuted}
\usepackage{tabularx}
\usepackage{gensymb}
\usepackage{multirow}
\usepackage{adjustbox}
\usepackage{array}
\usepackage{booktabs}
\usepackage{commath}
\usepackage{caption}
\usepackage{mathtools}
\usepackage{subcaption}
\usepackage[table]{xcolor} 

\usepackage{listings}
\lstdefinestyle{jsonbw}{
    language=,
    basicstyle=\small\ttfamily,
    breaklines=true,
    frame=lines,
    showstringspaces=false,
    emph={
        "model", "api_call", "payload", "instructions", "input",
        "role", "content", "type", "text", "image_url", "user",
        context_text, b64_img, b64_img1, b64_img2, category,
        object_name, scheme, dataset, min_frame_gap
    },
    emphstyle=\bfseries,
}

\input{preamble}
\definecolor{cvprblue}{rgb}{0.21,0.49,0.74}
\usepackage[pagebackref,breaklinks,colorlinks,allcolors=cvprblue]{hyperref}

\def\paperID{25} 
\def\confName{CVPR}
\def\confYear{2026}

\title{DietDelta: A Vision-Language Approach for Dietary Assessment via Before-and-After Images}

\author{
Gautham Vinod\quad
Siddeshwar Raghavan\quad
Bruce Coburn\quad
Fengqing Zhu
\\
\small{Purdue University, West Lafayette, Indiana, U.S.A.}\\
{\tt\small \{gvinod, raghav12, coburn6, zhu0\}@purdue.edu}
}
\begin{document}
\maketitle

\begin{abstract}
Accurate dietary assessment is critical for precision nutrition, yet most image-based methods rely on a single pre-consumption image and provide only coarse, meal-level estimates. These approaches cannot determine what was actually consumed and often require restrictive inputs such as depth sensing, multi-view imagery, or explicit segmentation. In this paper, we propose a simple vision–language framework for food-item-level nutritional analysis using paired \textit{before-and-after} eating images. Instead of relying on rigid segmentation masks, our method leverages natural language prompts to localize specific food items and estimate their weight directly from a single RGB image. We further estimate food consumption by predicting weight differences between paired images using a two-stage training strategy. We evaluate our method on three publicly available datasets and demonstrate consistent improvements over existing baselines, establishing a strong baseline for \textit{before-and-after} dietary image analysis.
\end{abstract}

\input{sec/01_introduction}
\input{sec/02_related_works}
\input{sec/03_methodology}
\input{sec/04_experimental_results}

\input{sec/05_conclusion}
{
    \small
    \bibliographystyle{ieeenat_fullname}
    \bibliography{main}
}


\end{document}

%% file: sec/01_introduction.tex
\section{Introduction}
Dietary habits play a key role in long-term health, strongly influencing the risk and management of chronic diseases such as obesity, type-2 diabetes, and cardiovascular diseases~\cite{adolph2024western}. While the importance of healthy eating is well established, accurately tracking daily nutritional intake remains a significant hurdle~\cite{nestle2025eat}. Typically, dietary assessment relies on individuals reporting their food consumption to a registered dietitian~\cite{subar2012automated}, which is time-consuming and prone to bias and inaccuracies. To address these limitations, Image-Based Dietary Assessment (IBDA) has emerged as a reliable alternative~\cite{konstantakopoulos2023automated}. Recent advances in deep learning have significantly improved the accuracy of IBDA methods by leveraging multimodal visual cues to recognize foods and infer their attributes~\cite{lo2020image}. However, precise portion size and nutritional estimation remain fundamentally challenging due to the ill-posed problem of recovering three-dimensional food geometry from two-dimensional images.

Current approaches to IBDA face three critical limitations. First, many methods rely on cumbersome and constrained inputs in addition to images, such as depth maps, multi-view sequences, 3D models, or specialized hardware, restricting their utility in real-world scenarios~\cite{shan2025depth, fujita2025mobile, Meyers_2015_ICCV, raju2022foodcam, Zeman2023EndToEnd, Vinod_2024_CVPR, subhi2018food, thames2021nutrition5k}. Second, many existing methods predict coarse, image-level nutritional profiles, overlooking the fine-grained, food-item-level analysis that is necessary for precision nutrition~\cite{kirk2021precision}. Third, and perhaps most importantly, current methods focus on \textit{before-eating images} (image of the food before it is eaten). This limitation stems from the intermediate components of IBDA methods, such as food segmentation, classification, and bounding-box detection, which are explicitly trained on intact, \textit{before-eating images}~\cite{Zeman2023EndToEnd, thames2021nutrition5k, zhao2024visual, vinod2026size, ma2024mfp3d}. By failing to analyze \textit{after-eating images}, these systems cannot accurately calculate the net consumption, rendering them ineffective for realistic dietary monitoring.

\begin{figure}[t!]
    \centering
    \includegraphics[width=\linewidth]{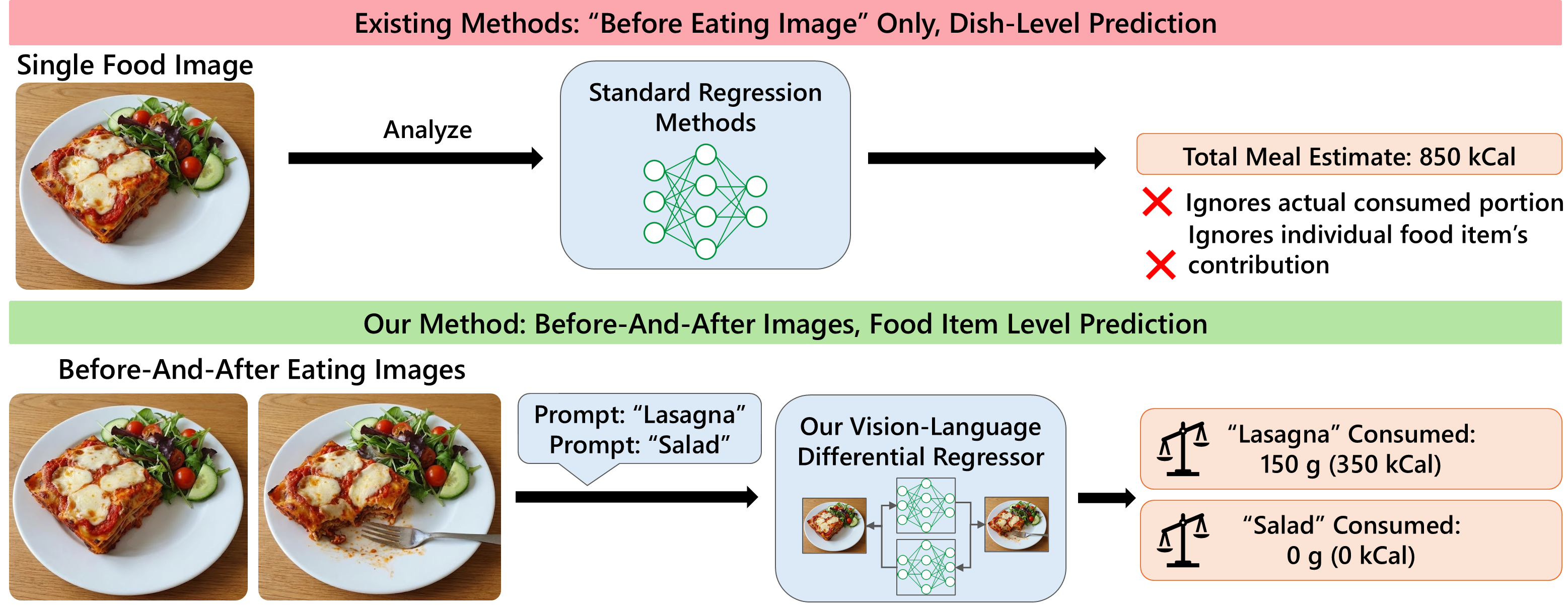}
    \caption{\textbf{Complete Consumption Analysis.} Our method distinguishes itself from existing approaches by analyzing the entire ``eating occasion'' and accounting for plate waste, to provide a precise nutritional breakdown of actual consumption.}
    \label{fig:intro}
    \vspace{-0.1cm}
\end{figure}

In this work, we address these challenges with a unified multimodal framework that operates on a single \textit{before-and-after image pair} and provides food-item-level nutritional estimates. 
Unlike prior methods that depend on complex segmentation pipelines, our approach utilizes natural language text prompts to localize specific food items and estimate their weight. This text-guided mechanism allows for flexible, per-item analysis without the need for rigid class definitions. By identifying the food item in paired images and utilizing a regression network to model the visual changes, our method directly calculates the difference in weight, providing a measure of actual intake rather than just the served portion size. Figure~\ref{fig:intro} summarizes the limitations of existing approaches and shows the major advantages posed by our method in the field of dietary assessment.

To overcome the limited availability of paired \textit{before} and \textit{after-eating} data, we introduce a two-stage training method consisting of \textit{1) Absolute Weight Estimation} and \textit{2) Weight Difference Estimation}. In the first stage, the model relies on large-scale datasets with weight annotations to learn to associate a food item text query with the appropriate visual regions and predict the absolute weight of a food item~\cite{thames2021nutrition5k, sanatbyek2025multitask}. In the second stage, the model is fine-tuned on a smaller dataset~\cite{coburn2025comprehensive} containing \textit{before-and-after image pairs} to learn the subtle changes in the visual signal for each food item, which corresponds to the consumed weight. 

We show that the performance of our \textit{absolute weight estimation} outperforms existing methods on publicly available datasets, and we extend this knowledge into \textit{weight difference estimation}, where we compare against baseline approaches such as direct image regression and large vision-language models. With this work, we aim to establish a solid benchmark for \textit{before-and-after image} analysis contributing to the advancement of the field of precision nutrition. The main contributions of our work can be summarized as:
\begin{itemize}
    \item We propose one of the first works to directly estimate food-item-level nutrition consumption from paired \textit{before-and-after eating} images.
    \item We introduce a text-guided attention mechanism that enables precise localization and weight estimation of individual food items using text queries, eliminating the need for complex pixel-wise segmentation.
    \item Our method is evaluated on three publicly available datasets, demonstrating superior performance over existing approaches and establishing a strong benchmark for future research in precision nutrition.
\end{itemize}

\begin{figure*}[htbp!]
    \centering
    \includegraphics[width=\linewidth]{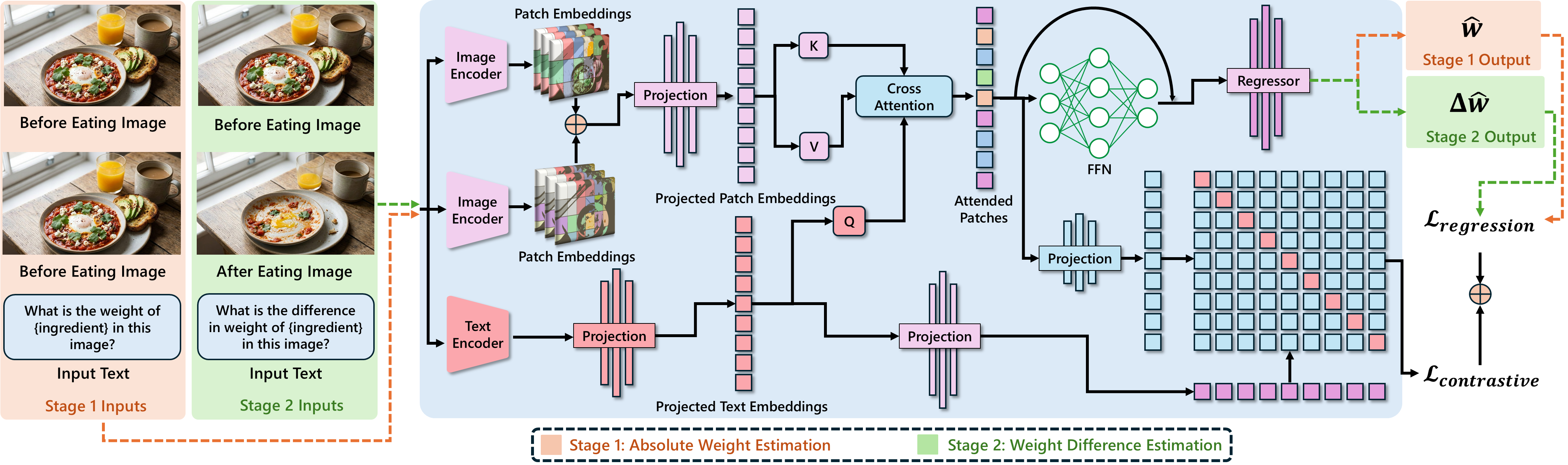}
    \caption{\textbf{Method Overview.} The two stage training strategy uses only the \textit{before} eating images in the \textit{Absolute Weight Estimation} stage to learn the patches related to the input prompt. This knowledge is used to finetune the model in the \textit{Weight Difference Estimation} stage to predict the weight difference of the food item in the text prompt. The text embeddings and image patch embeddings are fused to learn the most relevant image patches to the text prompt which is used to predict the absolute weight or weight difference depending on the stage.}
    \vspace{-0.1cm}
    \label{fig:method_overview}
\end{figure*}

%% file: sec/02_related_works.tex
\section{Related Works}

\noindent\textbf{Food Image Analysis.}
Computational food analysis presents unique challenges due to high intra-class variance (\textit{e.g., visual diversity within ``burgers''}) and inter-class similarity (\textit{e.g., visual overlap between ``apple pie'' and ``bread pudding''})~\cite{bossard2014food}. While prior work has explored fundamental tasks such as food classification and segmentation~\cite{tahir2021comprehensive, lo20200review}, portion estimation remains particularly challenging as it requires determining \textit{what is in the image}, \textit{where is it located}, and \textit{how much of it is present in the image}. This complexity is the primary obstacle to automated dietary assessment.

\noindent\textbf{Portion Size Estimation.}
The ill-posed nature of estimating volume from a single 2D image has led many researchers to rely on additional modalities~\cite{vinod2026food}. A common approach involves depth maps, utilized either for direct 3D reconstruction~\cite{graikos2020single, Lo2019DepthEstimation}, voxel-based modeling~\cite{Zeman2023EndToEnd, fang2016comparison, myers2015im2calories}, or as an input to a network that fuses RGB and depth information for portion predictions~\cite{thames2021nutrition5k, Feng2025RGBD, zhang2025ingredients}. However, these methods typically require ground-truth depth during training, and sometimes even inference, which restricts their practicality in real-world settings where such data is unavailable.
Alternative approaches estimate portion size via point cloud reconstruction from multiple images~\cite{Puri2009Recognition, Dehais2017TwoView}. While effective, these methods impose a heavy burden on the user (requiring image capture from multiple angles) and necessitate a physical reference object (fiducial marker) for scale calibration~\cite{vinod20243dobject}. In contrast, our approach eliminates these hardware and capture constraints, performing weight estimation from a single RGB image guided by natural language prompts.

\noindent\textbf{Consumption and Plate Waste Analysis.} Current systems largely ignore plate waste, analyzing only \textit{before-eating images} and implicitly assuming complete consumption. Analysis of the \textit{after-eating image} is typically avoided because half-eaten, mixed leftovers confuse existing models. While some methods compare leftovers against rigid standardized shapes~\cite{martin2014measuring}, these fail when food structure is lost. We address this by directly quantifying the weight difference between \textit{before} and \textit{after} states, providing a precise record of actual intake.

%% file: sec/03_methodology.tex
\section{Methodology}

Our method leverages the patch-based architecture of Vision Transformers (ViT)~\cite{dosovitskiy2020image} to achieve fine-grained, text-guided food-item localization. Unlike standard multimodal models like CLIP~\cite{radford2021learning} that typically align visual and textual representations at a global image level, our method adapts this mechanism to operate at the patch level (Figure~\ref{fig:method_overview}). By querying these patches with text features, we identify the specific spatial regions associated with a food item and then predict both absolute weight and weight differences.

\subsection{Obtaining the Embeddings}
To enable text-guided localization, we first extract representations for both the visual and textual inputs. The textual input for food-item localization would be generated by an upstream food classifier or localizer. Our approach aims at tackling the fundamentally harder problem of using the classification result for accurate size estimation. Hence, we naively use the ground-truth food classes for textual prompts in our method. We select the text prompts to match the objective of the stage. For \textit{Stage 1 - Absolute Weight Estimation}, we construct the text prompt as \textit{``What is the weight of the [FOOD-ITEM] in this image?''}. For \textit{Stage 2 - Weight Difference Estimation}, the text prompt used to learn the visual differences of a food item in the image is \textit{``What is the difference in weight of the [FOOD-ITEM] in these images?''}.

We adopt the ViT-L/14 variant of CLIP~\cite{radford2021learning} as our backbone. 
The image encoder $\mathcal{E}_{\textit{vision}}$ divides each \textit{before} image ($I_{\textit{before}}$) and \textit{after} image $(I_{\textit{after}})$ into $N=576$ patches pf size  $14 \times 14$ to produce fine-grained patch embeddings. 
Simultaneously, the text query $T_{\textit{input}}$ is encoded via $\mathcal{E}_{\textit{text}}$ to provide a semantic reference vector $\mathbf{t}$. This yields the following feature representations:
\begin{gather*}
    \mathbf{F}_{k} = \mathcal{E}_{\textit{vision}}(I_{k}) \in \mathbb{R}^{N \times D_I}, \quad \text{for } k \in \{\textit{before}, \textit{after}\} \\
    \mathbf{t} = \mathcal{E}_{\textit{text}}(T_{\textit{input}}) \in \mathbb{R}^{1 \times D_T}
\end{gather*}
\noindent where $D_I$ and $D_T$ denote the embedding dimensions of the image patches and text, respectively. For Stage 1, $I_{\textit{before}}$ and $I_{\textit{after}}$ are the same ``\textit{before-eating}'' image to train the regressor for ``\textit{absolute weight estimation.}'' This helps us prevent architecture changes from Stage 1 to Stage 2.

\subsection{Cross-Attention and Regression}
We then pass the extracted image features ($\mathbf{F}_{\textit{before}},\mathbf{F}_{\textit{after}}$) and text feature ($\mathbf{t}$) through MLP layers to project them into a combined, lower-dimensional embedding space. This reduction is essential to 1) ensure the dimensionality of the features spaces are the same for the upcoming cross-attention mechanism, 2) learn a task-specific alignment between the visual and textual modalities, and 3) enable the model to efficiently compute similarities and seamlessly merge patch information from both input images.
\vspace{-0.1cm}
\begin{gather*}
    \mathbf{H}_{\textit{img}} = \phi_{\textit{img}}\left( [\mathbf{F}_{\textit{before}}  \oplus \mathbf{F}_{\textit{after}}] \right) \\
    \mathbf{q}_{\textit{text}} = \phi_{\textit{text}}(\mathbf{t})
\end{gather*}
\vspace{-0.1cm}
\noindent where $\oplus$ denotes concatenation and $\phi$ are the projection MLPs.

A key component of our framework is the modified cross-attention mechanism~\cite{vaswani2017attention}. In \textit{Stage 1 - Absolute Weight Estimation}, the cross-attention module semantically aligns image patches with the text query. The projected text embeddings $\mathbf{q}_{\textit{text}}$ acts as the Query (Q), while the image patch embeddings $\mathbf{H}_{\textit{img}}$ serve as both Keys (K) and Values (V). Consequently, the module outputs a weighted aggregation of the visual features, where the attention weights are determined by the relevance of each patch to the text description (query). This is obtained by attending the $(Q, K, V)$ vectors using:
\vspace{-0.1cm}
\begin{gather*} 
Q = \mathbf{q}_{\textit{text}}, \quad K = \mathbf{H}_{\textit{img}}, \quad V = \mathbf{H}_{\textit{img}} \\ 
\mathbf{z}_{\textit{attn}} = \text{softmax}\left( \frac{Q K^\top}{\sqrt{d_k}} \right) V 
\end{gather*}
where $d_k$ is the dimension of each vector, and the output $\mathbf{z}_{\textit{attn}}$ are the attended patch embeddings with respect to the text query. We show via qualitative results of the weighted patches that our method learns to accurately identify the relevant patches for each food-item mentioned in the text query, as seen in Figure~\ref{fig:heatmaps}. 
In \textit{Stage 2 - Weight Difference Estimation}, cross-attention weights reflect the difference in detected food item quantities between the two images. Across both stages, the attended patches provide a strong signal for weight prediction. This signal is processed by a Feed-Forward Network (FFN), which adds a non-linear transformation to learn complex patterns in the signal and helps convert the context-aware attended patches into a richer feature-specific representation for the regression task. 

We also use the attended patches as a residual input to ensure better information flow across the model for more stable learning. 
The last regression head $\mathcal{R}$ projects this enriched representation from the FFN and uses it to predict a weight difference $\Delta \hat{w}$ of the food item: 
\begin{gather*}
    \mathbf{h}_{\textit{res}} = \mathcal{F}_{\text{FFN}}(\mathbf{z}_{\textit{attn}}) + \mathbf{z}_{\textit{attn}} \\
    \Delta \hat{w} = \mathcal{R}(\mathbf{h}_{\textit{res}}) \in \mathbb{R}^{1 \times 1}
\end{gather*}
\noindent where $\mathcal{F}_{\text{FFN}}$ denotes the FFN with non-linear activations.

\subsection{Objective Function}
To train our model, we employ a multi-task objective function that simultaneously optimizes for accurate weight difference estimation and semantic alignment between the attended image features and the input text. The total loss is defined as a weighted sum of a regression loss and a contrastive alignment loss:

\begin{equation} \label{eq:loss}
\mathcal{L}_{\textit{total}} = \lambda_{\textit{reg}}\mathcal{L}_{\textit{reg}} + \lambda_{\textit{cont}}\mathcal{L}_{\textit{contrastive}}
\end{equation}
\noindent where $\lambda_{\textit{reg}}$ and $\lambda_{\textit{cont}}$ are hyperparameters governing the contribution of each task. 

The primary goal of our network is to minimize the error in weight difference estimation.
We utilize the $L_1$ loss between the estimates and the targets for our regression loss. During \textit{Stage 1 - Absolute Weight Estimation}, the loss between the predicted weight $\hat{w}$ and the ground-truth weight $w$ is formulated as:

\begin{equation}
\mathcal{L}_{\textit{reg}} = \frac{1}{B} \sum_{i=1}^{B} | w_i - \hat{w}_i |
\end{equation}

Further, for \textit{Stage 2 - Weight Difference Estimation}, the predicted weight difference $\Delta \hat{w}$ and the ground truth weight difference $\Delta w$ are used in the regression loss as:
\begin{equation}
\mathcal{L}_{\textit{reg}} = \frac{1}{B} \sum_{i=1}^{B} | \Delta w_i - \Delta \hat{w}_i |
\end{equation}
\noindent where $B$ denotes the batch size.

Further, to ensure that the cross-attention mechanism effectively highlights semantically relevant patches, we explicitly align the attended image representation $\mathbf{z}_{\textit{attn}}$ with the text embedding $\mathbf{t}$. 
We apply a contrastive loss (InfoNCE~\cite{radford2021learning}) $\mathcal{L}_{\textit{contrastive}}$ to maximize the similarity between matched image patch-text pairs while suppressing the similarity of unmatched pairs within the batch. 


%% file: sec/04_experimental_results.tex
\section{Experimental Results}
\begin{table*}[!ht]
    \footnotesize
    \centering
    \setlength{\tabcolsep}{1pt} 
    
    \begin{tabularx}{\textwidth}{l >{\centering\arraybackslash}X >{\centering\arraybackslash}X >{\centering\arraybackslash}X >{\centering\arraybackslash}X >{\centering\arraybackslash}X >{\centering\arraybackslash}X >{\centering\arraybackslash}X}
    \toprule
    \multirow{2}{*}{\textbf{Method}} & \multicolumn{2}{c}{\textbf{Nutrition5k}} & \multicolumn{2}{c}{\textbf{FPB}} & \multicolumn{2}{c}{\textbf{ACE-TADA}} & \multirow{2}{\linewidth}{\textbf{Mean PMAE (\%)}}\\ 
    \cmidrule(lr){2-3} \cmidrule(lr){4-5} \cmidrule(lr){6-7}
    
    &
    \multicolumn{1}{c}{\textbf{MAE (g)}} & \multicolumn{1}{c}{\textbf{PMAE (\%)}} & 
    \multicolumn{1}{c}{\textbf{MAE (g)}} & \multicolumn{1}{c}{\textbf{PMAE (\%)}} & 
    \multicolumn{1}{c}{\textbf{MAE (g)}} & \multicolumn{1}{c}{\textbf{PMAE (\%)}} & \\
    \midrule
    Baseline & 124.6 & 60.2 & 137.33 & 55.30 & 237.91 & 28.86 & 48.12 \\
    
    RGB*~\cite{thames2021nutrition5k} & 41.56 & 20.94 & 84.74 & 34.12 & 356.78 & 43.26 & 32.77 \\
    
    RGB-D*~\cite{thames2021nutrition5k} & 71.17 & 35.85 & 94.98$^\dagger$ & 38.25$^\dagger$ & 292.15$^\dagger$ & 35.64$^\dagger$ & 36.58 \\
    
    Yolo-v12S Predictor~\cite{sanatbyek2025multitask} & No Bbox & No Bbox & 90.95 & 44.60 & No Bbox & No Bbox & 44.60 \\
    
    Swin Nutrition*~\cite{shao2022rapid} & 74.28 & 37.42 & 165.65 & 66.70 & 234.72 & 28.46 & 44.19 \\
    
    Closed VLM (\emph{Gemini 2.5 Pro~\cite{gemini_team_2025_gemini25}}) & 74.87 & 44.76 & 65.76 & 40.09 & 176.12 & 24.76 & 36.54 \\
    
    Open VLM (\emph{Gemma 3 27B~\cite{gemma_team_2025_gemma3}}) & 71.96 &74.88 & 102.49 & 45.87 & 223.13 & 26.21 & 48.99 \\
    
    \textbf{DietDelta (Ours)} & \textbf{35.10} & \textbf{17.68} &  \textbf{38.29}  &  \textbf{15.25} &  \textbf{85.27} &  \textbf{10.34} & \textbf{14.42} \\
    \bottomrule
    \end{tabularx}
    
    \caption{\small\textbf{Absolute Weight Estimation Comparison.} Quantitative results comparing DietDelta (Ours) with existing deep learning and VLM-based methods. Our approach utilizes both image and text modalities to predict food-item weights, achieving a substantial reduction in error compared to other methods with a Mean PMAE of 14.42\%. 
    \textit{* indicates methods reimplemented via DeepCode~\cite{li2025deepcode}. $^\dagger$ indicates depth maps obtained using Metric3D v2~\cite{hu2024metric3d} (no ground-truth depth available).}}
    \label{tab:weight_estimation_results}
    \vspace{-0.2cm}
\end{table*}

\subsection{Experimental Setup}

\noindent\textbf{Datasets.} To rigorously evaluate our two-stage framework, we utilize three publicly available dietary datasets, each serving a specific role in our training pipeline:
\begin{itemize}
    \item \textbf{Nutrition5k~\cite{thames2021nutrition5k}:} This dataset consists of 2,758 training and 507 testing RGB-D images of complex, multi-ingredient meals. As it only contains pre-consumption imagery, we utilize the RGB images and their corresponding ingredient-level weight annotations exclusively to train the \textit{Absolute Weight Estimation} (Stage 1).
    \item \textbf{Food Portion Benchmark (FPB)~\cite{sanatbyek2025multitask}:} Comprising 11,718 images, this dataset provides rich bounding box and weight annotations. We leverage this large-scale data to further strengthen and extend our Stage 1 training, teaching the model to accurately associate text prompts with specific visual food regions.
    \item \textbf{ACE-TADA~\cite{coburn2025comprehensive}:} To evaluate our core contribution of consumption tracking, we utilize the 806 paired ``Before-and-After'' eating images from this dataset. We apply a standard 80:20 train-test split and use this data exclusively for fine-tuning and evaluating the \textit{Weight Difference Estimation} (Stage 2).
\end{itemize}

\noindent\textbf{Implementation Details.} Our framework is implemented in PyTorch and trained end-to-end on a single NVIDIA A40 GPU. For our feature extraction backbone, we employ pre-trained CLIP (ViT-L/14@336px)~\cite{radford2021learning} models for both the image and text modalities. Crucially, the CLIP encoders are kept completely frozen during training. This deliberate design choice preserves the rich, contrastive semantic alignment learned during large-scale pre-training and acts as a strong regularizer against overfitting on our relatively small dataset of paired before-and-after images. 

During training, we optimize the network using the AdamW optimizer with a base learning rate of $1e^{-4}$ and a weight decay of $1e^{-2}$. The learning rate is decayed using a Cosine Annealing schedule over 150 epochs. The total loss (Eq.~\ref{eq:loss}) is a weighted sum of the regression loss and the cross-attention alignment loss, with hyperparameters empirically set to $\lambda_{reg}=1.0$ and $\lambda_{cont}=0.2$ to carefully balance precise weight estimation with robust text-image feature matching.

\noindent\textbf{Evaluation Metrics.} We evaluate the predictive performance of DietDelta using Mean Absolute Error (MAE) and Percentage Mean Absolute Error (PMAE). PMAE provides a normalized view of the error by scaling the MAE by the mean ground-truth weights of the test set, making it highly effective for comparing performance across datasets with varying portion sizes.

For \textit{Stage 1 - Absolute Weight Estimation}, the metrics are defined as:
\vspace{-0.2cm}
\begin{align}
    \text{MAE} = \frac{1}{N} \sum_{i=1}^N | w_i - \hat{w}_i|  \\
    \text{PMAE} =  \frac{1}{N} \sum_{i=1}^N \frac{| w_i - \hat{w}_i|}{\bar{w}}
\end{align}
\vspace{-0.2cm}

\noindent where $N$ is the total number of evaluated dishes, $w_i$ and $\hat{w}_i$ are the ground-truth and predicted absolute weights respectively, and $\bar{w}$ is the mean ground-truth weight.

For \textit{Stage 2 - Weight Difference Estimation}, we compute the metrics identically, but replace the absolute weights with the differential mass ($\Delta w_i = w_{i,\text{before}} - w_{i,\text{after}}$). Thus, the model is evaluated strictly on its ability to quantify the actual consumed amount.

\noindent\textbf{Item-to-Dish Aggregation.} While DietDelta's text-guided attention inherently performs instance-level regression for individual food items, many existing baseline methods lack this fine-grained capability and predict total meal weight holistically. To ensure a fair and direct quantitative comparison against these baselines in our results, we aggregate (sum) our item-level weight predictions to calculate the total dish-level error. This aggregation strategy is applied to all reported MAE and PMAE values in our tables.

\subsection{VLM Setup, Prompt Engineering, and Structured Reasoning}
To evaluate the efficacy of \textit{DietDelta} against state-of-the-art generalist models, we conduct experiments using Gemini 2.5 Pro~\cite{gemini_team_2025_gemini25} and Gemma 3 27B~\cite{gemma_team_2025_gemma3}. For Gemini 2.5 Pro, we utilize the \texttt{google-genai} library with default temperature settings. Gemma 3 27B is deployed in native \texttt{bfloat16} precision on an NVIDIA H100 GPU to ensure numerical stability without quantization artifacts. We utilize greedy decoding (max 512 tokens) to maintain deterministic outputs for portion estimation.

The VLM baseline weight estimation prompts are designed to (1) instruct the models to estimate weight and (2) output the predictions in a structured manner. The VLMs were instructed to provide weight predictions either using a single meal image or before-and-after meal image pairs. Across these two types of experiments, the same ``structured output" was requested at the end of the prompt. These portions of the prompts are included below. It should be noted that ingredient names were provided within each prompt to align with how DietDelta similarly provides class-based text guidance. Prompts are provided for transparency and reproducibility.

\paragraph{Single Meal Image Weight Estimation Prompt}
Listing \ref{lst:mono_prompt} provides the single-image prompt asking the model to estimate the weight of the meal provided in the image. \texttt{ing\_list} corresponds to a list of ingredients present in the meal image, determined by the ground truth labels provided within each dataset.

\begin{lstlisting}[style=jsonbw, caption={Single Meal Image Weight Estimation Prompt}, label={lst:mono_prompt}]
"You are a nutrition expert analyzing this meal image. Estimate the weight in grams for these ingredients:\n{ing_list}"
\end{lstlisting}

\paragraph{Before-and-After Meal Image Prompts}

We ask the model to predict weight when given before-and-after meal image pairs based on two different configurations: (1) \emph{Predicted Difference} - We directly ask the models to estimate the consumed weight of the meal when provided the before-and-after meal image pair directly, (2) \emph{Difference of Predictions} - We separately provide the before and after meal images, asking the model to estimate the weight of each meal image, then calculating the difference between the two estimates. The prompt used for Predicted Difference is provided in Listing \ref{lst:pred_diff} and for Difference of Predictions in Listings \ref{lst:diff_of_pred_before} and \ref{lst:diff_of_pred_after} for the before and after images, respectively. Like in Listing \ref{lst:mono_prompt}, \texttt{ing\_list} corresponds to a list of ingredients present in the meal image(s), determined by the ground truth labels provided within each dataset.

\begin{lstlisting}[style=jsonbw, caption={Predicted Difference Prompt}, label={lst:pred_diff}]
"You are a nutrition expert. Analyze these two images. "
Image 1 is the meal Before eating. Image 2 is the meal After eating.
Identify the following ingredients: {ing_list}.
Estimate the CONSUMED weight (mass eaten) in grams for each ingredient based on the difference between the images.
Example: {{\"Apple\": 50.5, \"Bread\": 20.0}}"
\end{lstlisting}

\begin{lstlisting}[style=jsonbw, caption={Difference of Predictions Prompt (Before Image)}, label={lst:diff_of_pred_before}]
"You are a nutrition expert. Analyze this image of a meal.
Estimate the total weight (in grams) PRESENT in the image for these ingredients: {ing_list}."
\end{lstlisting}

\begin{lstlisting}[style=jsonbw, caption={Difference of Predictions Prompt (After Image)}, label={lst:diff_of_pred_after}]
"You are a nutrition expert. Analyze this image of leftovers/after meal.
Estimate the remaining weight (in grams) PRESENT in the image for these ingredients: {ing_list}.
If an ingredient is completely gone, the weight is 0.
\end{lstlisting}

\paragraph{Structured Output Prompt}

Listing \ref{lst:struc_out_prompt} provides the part of the prompt responsible for outputting a structured output (leading to readily parseable results) and ensuring that the model does provide a guess for every provided ingredient. This ``sub-prompt" is appended to every weight estimation prompt.

\begin{lstlisting}[style=jsonbw, caption={Structured Output Prompt (appended to the weight estimation prompts}, label={lst:struc_out_prompt}]
"RULES:
1. Output ONLY a valid JSON object.
2. Keys must be the exact ingredient names listed.
3. Provide a best-guess estimate in grams.
4. Example: {{\"Rice\": 150.0, \"Chicken\": 85.0}}"
\end{lstlisting}

\subsection{Results}

\textbf{Metric Aggregation for Fair Comparison.} A key contribution of DietDelta is its ability to perform precise, instance-level regression for individual food items based on text prompts (Figure~\ref{fig:heatmaps}). However, existing baselines (e.g., RGB predictors and standard VLMs) are typically designed to predict the total meal weight and lack the capability to isolate specific ingredients without explicit segmentation masks. Therefore, to ensure a fair and direct quantitative comparison in Table~\ref{tab:weight_estimation_results} and Table~\ref{tab:weight_difference_estimation}, we aggregate (sum) our item-level predictions per dish. This demonstrates that even when evaluated at the macroscopic meal level, our item-specific reasoning yields superior accuracy.

\noindent \textbf{Absolute Weight Estimation.} We compare DietDelta against a comprehensive suite of baselines, ranging from traditional deep learning regressors (RGB and RGB-D~\cite{thames2021nutrition5k}, Swin Nutrition~\cite{shao2022rapid}),  to open (Gemma 3 27B~\cite{gemma_team_2025_gemma3}) and closed source (Gemini 2.5 Pro~\cite{gemini_team_2025_gemini25}) large Vision-Language Models (VLMs).  The quantitative results on the Nutrition5k, FPB, and ACE-TADA datasets are summarized in Table~\ref{tab:weight_estimation_results}.

While RGB~\cite{thames2021nutrition5k} achieves reasonable performance on Nutrition5k, these methods generally struggle to generalize across datasets. Gemini 2.5 Pro demonstrates competitive capabilities (36.54\% Mean PMAE), it suffers from high latency (25.31s per inference), which is impractical for real-time applications. Conversely, the smaller and faster Gemma 3 model lacks the visual reasoning capacity for accurate weight regression.

DietDelta consistently outperforms all baselines, achieving a \textbf{Mean PMAE of 14.42\%}. By effectively fusing semantic text priors with visual features, our method reduces the error by more than 50\% compared to the strongest baseline (RGB).

\begin{figure}[htpb]
    \centering
    \begin{subfigure}[b]{\linewidth}
        \centering
        \includegraphics[width=\linewidth]{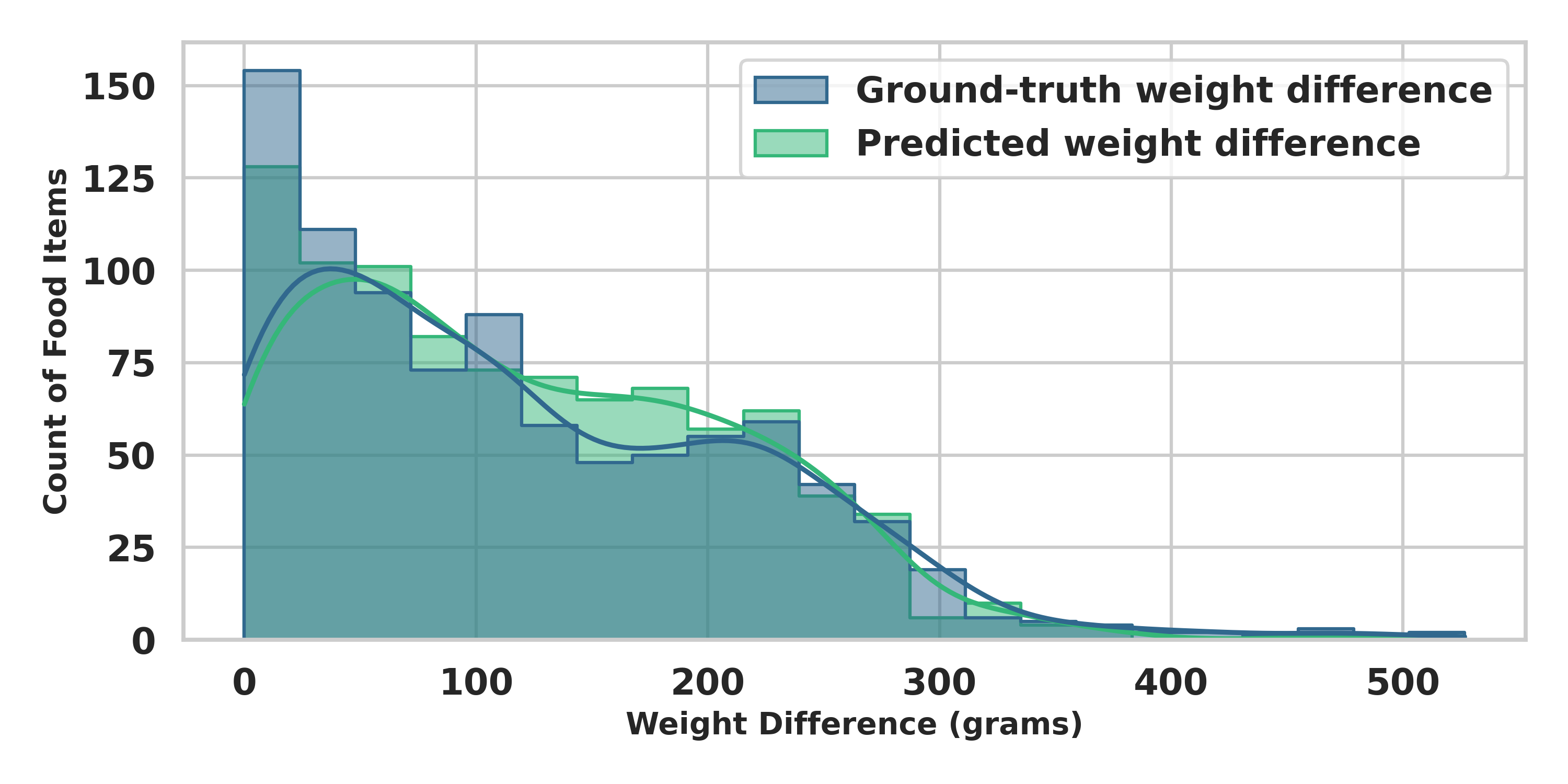} 
        \caption{Distribution of Predicted and Ground Truth Weight Differences}
        \label{fig:histogram}
    \end{subfigure}

    \begin{subfigure}[b]{\linewidth}
        \centering
        \includegraphics[width=\linewidth]{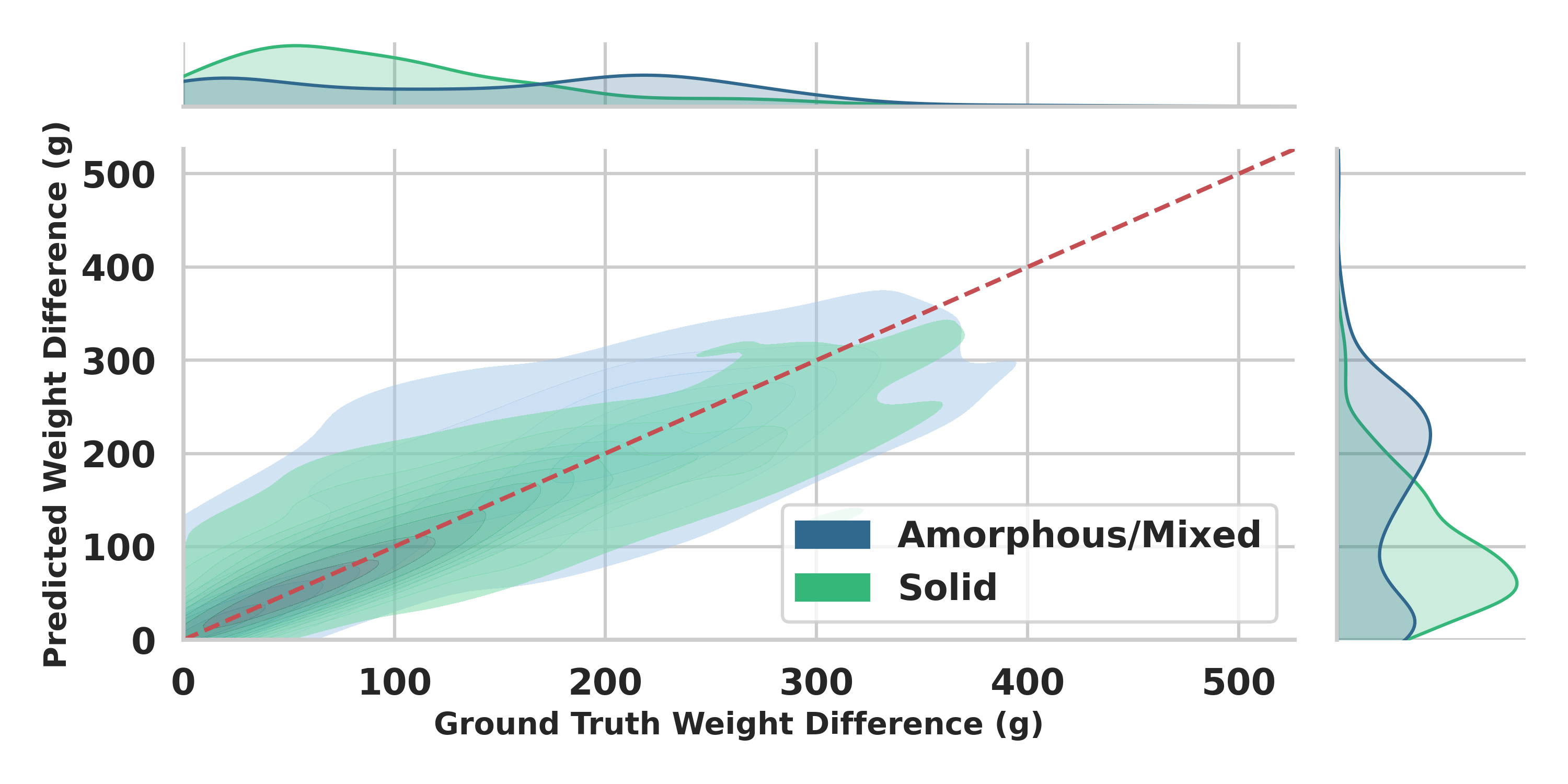} 
        \caption{Joint Density of Predicted vs. Ground Truth Weight Differences by Food Structure}
        \label{fig:kde}
    \end{subfigure}
    \caption{\textbf{Predicted and Ground Truth Weight Difference Analysis.} (a) Frequency distribution of weight differences showing strong overlap between predictions and ground truth. (b) Bivariate density plot of predictions vs. ground truth, stratified by food structure. The alignment along the $y=x$ diagonal across both Solid and Amorphous types confirms the model's generalization capability.}
    \label{fig:grid}
\end{figure}

\noindent \textbf{Weight Difference Estimation.} To evaluate the capability of our model in dietary intake monitoring, we evaluate performance on \textit{before-and-after} image pairs from the ACE-TADA dataset. We compare our approach against three distinct strategies, shown in Table~\ref{tab:weight_difference_estimation}:

\noindent 1) RGB Difference Predictor: An extension of the RGB method~\cite{thames2021nutrition5k} where a model trained on the \textit{before-eating} images and another trained on \textit{after-eating} images make independent predictions and their difference is calculated.

\noindent 2) Closed VLM (Predicted Difference): The VLM is provided both images simultaneously and prompted to estimate the consumed amount directly.

\noindent 3) Closed VLM (Difference of Predictions): The VLM estimates the absolute weight of the food in the ``Before'' and ``After'' images independently, and the difference is calculated.

\begin{table}[h!]
    \centering
    \begin{adjustbox}{max width=\linewidth}
    \begin{tabular}{l|c c}
        \textbf{Method} & \textbf{MAE (g)} & \textbf{PMAE (\%)} \\
        \midrule
         Baseline & 190.24 & 27.51 \\
         RGB Difference Predictor &  374.85 & 54.21 \\
         Closed VLM (\emph{Predicted Difference}) & 223.14 & 38.90 \\
         Closed VLM (\emph{Difference of Predictions}) & 240.74 & 38.95 \\
         \textbf{DietDelta (Ours)} & \textbf{99.09}  & \textbf{14.17} 
    \end{tabular}
    \end{adjustbox}
    \caption{\small\textbf{Weight Difference Estimation Results.} We compare our proposed method against baseline and VLM approaches on \textit{before-and-after-eating} images. DietDelta (Ours) yields the lowest error rates across both metrics.}
    \label{tab:weight_difference_estimation}
\end{table}

As shown in Table~\ref{tab:weight_difference_estimation}, traditional approaches struggle significantly with the complex temporal reasoning required for difference estimation. The \textit{RGB Difference Predictor} fails catastrophically (54.21\% PMAE), as processing the ``Before'' and ``After'' images independently compounds geometric estimation errors. While the VLM-based approaches (\textit{Predicted Difference} and \textit{Difference of Predictions}) perform better, they still suffer from high error rates (approx. 39\% PMAE). This highlights a fundamental limitation in standard VLMs: they rely on holistic, generative reasoning that struggles to maintain precise, metric-level consistency across two distinct visual states.

In contrast, DietDelta achieves a \textbf{PMAE of 14.17\%}. This substantial improvement validates our multi-modal architecture. Unlike the generative nature of large VLMs, DietDelta's patch-level cross-attention is mathematically designed to compute spatial feature differences. By explicitly learning to correlate the textual prior with the \textit{missing} or \textit{altered} visual patches between the ``Before'' and ``After'' states, DietDelta effectively models the visual change directly. This allows the network to bypass the compounded errors of independent estimation and directly regress the precise consumption differential.

Further, we analyze the reliability of our predictions by visualizing the error distribution in Figure~\ref{fig:grid}. The histogram in Figure~\ref{fig:histogram} shows a strong overlap between the ground-truth (blue) and predicted (green) weight differences, indicating our model accurately captures the dataset distribution without systematic bias. 
Crucially, Figure~\ref{fig:kde} demonstrates our model's performance across food structures. Intuitively, amorphous foods (e.g., mashed potatoes, curry) are harder to estimate than distinct solids (e.g., apples, bread) due to undefined geometries. However, DietDelta maintains high accuracy (clustering along the diagonal) for both \textbf{Solid} and \textbf{Amorphous/Mixed} categories, validating the efficacy of text-guided attention in handling complex food shapes.

\subsection{Ablation Studies}
We analyze the effect of different components of our method and how our method's performance is affected. All the ablation experiments are performed on the Nutrition5k dataset for absolute weight estimation. 

\subsubsection{Effect of Text and Image Fusion}
A core hypothesis of this work is that text provides a critical semantic ``anchor'' which is validated in Table~\ref{tab:ablation_modality}. 
Using \textbf{Image Only} features results in a high MAE of 97.79g, as the model struggles to infer weight from visual cues alone. Interestingly, the \textbf{Text Only} baseline performs better (73.68g MAE) than vision-only, likely because it learns the average statistical weight of specific food classes. However, the fusion of both modalities in DietDelta yields a drastic improvement. This confirms that the modalities are complementary: text provides localization and class-specific priors, while the vision provides instance-specific size information.

\begin{table}[h!]
    \centering
    \begin{tabular}{c | c c}
        \textbf{Input Modality} & \textbf{MAE (g)} & \textbf{PMAE (\%)} \\
        \midrule
         Image Only & 97.79 & 49.26 \\
         Text Only & 73.68 & 37.12 \\
         \textbf{Image + Text (Ours)} & \textbf{35.10} & \textbf{17.68}\\
    \end{tabular}
    \caption{\small \textbf{Impact of Cross-Modal Feature Fusion.} Analysis of single-modality versus multi-modality performance shows that while text features alone provide strong cues, the fusion of image and text results in a drastic reduction in error.}
    \label{tab:ablation_modality}
\end{table}

\subsubsection{Effect of Encoder}
We investigate the impact of different pre-trained backbones on performance in Table~\ref{tab:ablation_encoder}. We test combinations of \textbf{CLIP}~\cite{radford2021learning}, \textbf{SigLIP2}~\cite{tschannen2025siglip}, and the domain-specific \textbf{RecipeBERT}~\cite{Mereddy2024RecipeBERT}.
Counter-intuitively, simply using the SigLIP2 does not yield the best results as compared to the symmetric \textbf{CLIP-CLIP} configuration, likely due to the feature space being more suited to our task. We observe that ``hybrid'' configurations (e.g., CLIP Image + RecipeBERT Text) perform significantly worse (53.54g MAE). This suggests that the alignment of the feature spaces derived from the original contrastive pre-training of the image and text encoders is more critical for our cross-attention mechanism than the raw capacity of the individual encoders.

\begin{table}[h!]
    \centering
    \begin{adjustbox}{max width=\linewidth}
    \begin{tabular}{c c | c c}
        \textbf{Image Encoder} & \textbf{Text Encoder} & \textbf{MAE (g)} & \textbf{PMAE (\%)} \\
        \midrule
        SigLIP2~\cite{tschannen2025siglip} & SigLIP2~\cite{tschannen2025siglip} & 37.69 & 18.99 \\
        SigLIP2~\cite{tschannen2025siglip} & CLIP~\cite{radford2021learning} & 43.28 & 21.80 \\
        CLIP~\cite{radford2021learning} & SigLIP2~\cite{tschannen2025siglip} & 47.26 & 23.81 \\
        CLIP~\cite{radford2021learning} & RecipeBERT~\cite{Mereddy2024RecipeBERT} & 53.54 & 26.97 \\
        \textbf{CLIP~\cite{radford2021learning}} & \textbf{CLIP~\cite{radford2021learning}} & \textbf{35.10} & \textbf{17.68} \\
    \end{tabular}
    \end{adjustbox}
    \caption{\small \textbf{Effect of Encoder Selection.} Performance analysis of different backbone combinations. We observe that the standard CLIP image and text encoders provide the most ideal representations for our task.}
    \label{tab:ablation_encoder}
\end{table}


\subsection{Qualitative Analysis}

\begin{figure}[htpb]
    \centering
    \includegraphics[width=\linewidth]{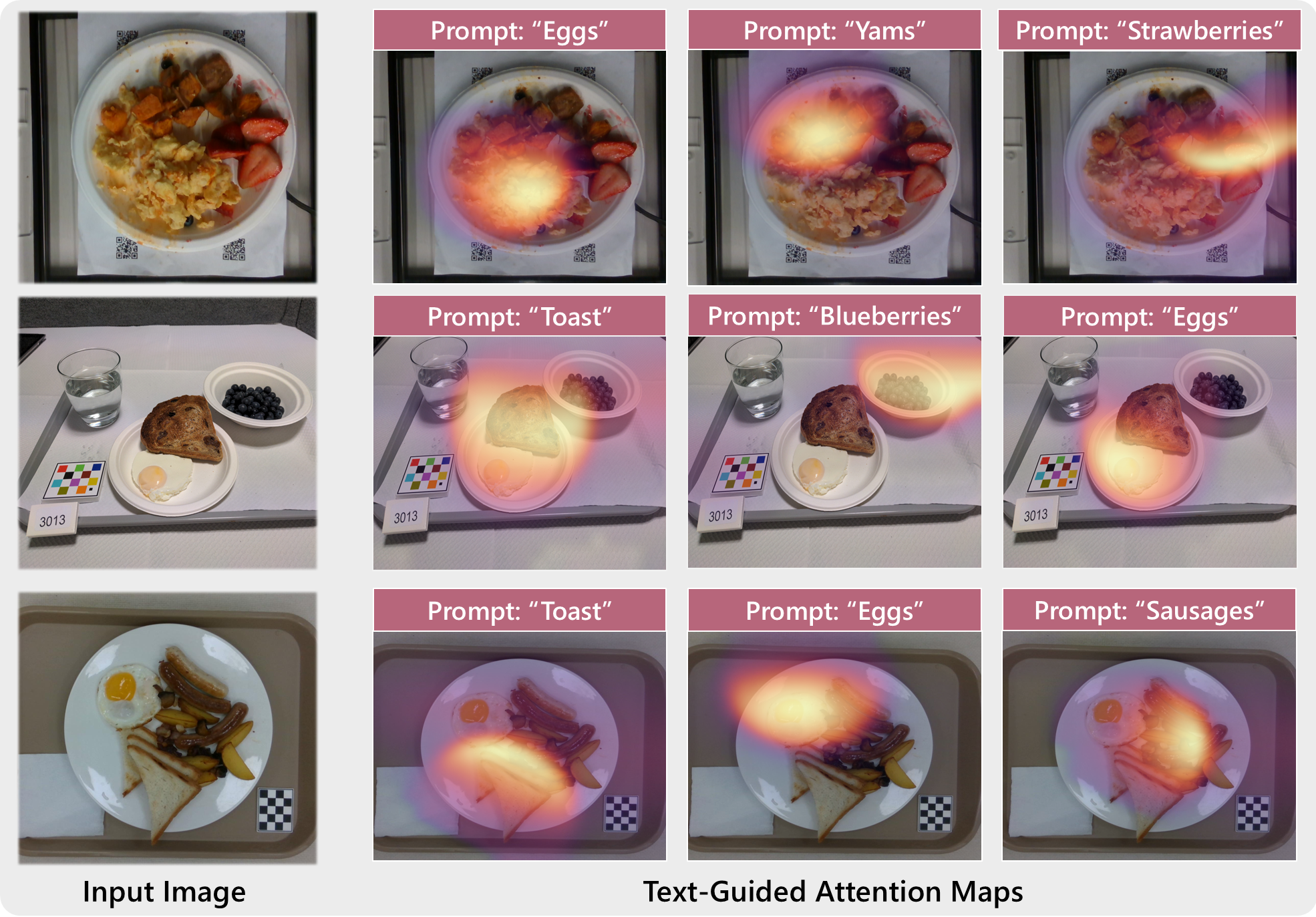}
    \caption{\textbf{Qualitative Results.} Images from the different datasets are analyzed with corresponding text prompts to show the activation of the images patches via the cross-attention mechanism.}
    \label{fig:heatmaps}
\end{figure}

The text prompt serves as a semantic anchor, guiding the model's attention to the specific region of interest within the complex scene of a meal. To validate that our model successfully learns this text-to-visual correspondence without explicit bounding box supervision, we visualize the cross-attention weights assigned to the image patches.

Figure~\ref{fig:heatmaps} illustrates these attention heatmaps across various text prompts. Notably, the model demonstrates high spatial precision: when prompted with specific ingredients, the high-activation regions (red/yellow) tightly cluster around the corresponding food items, effectively ignoring irrelevant background elements or adjacent foods. This qualitative evidence confirms that DietDelta is not merely relying on holistic image statistics (a common "shortcut" in standard regressors), but is actively isolating and analyzing the geometric features of the requested item to perform precise weight estimation.

%% file: sec/05_conclusion.tex
\section{Conclusion}
In this work, we presented an innovative method for precise, food-item-level dietary assessment capable of estimating both absolute food weight and consumed amounts from \textit{before-and-after} image pairs. Our core contribution lies in the effective application of cross-attention mechanisms, which allow natural language prompts to act as semantic anchors. This helps us achieve superior performance with a Mean PMAE of 14.42\% across three publicly available datasets. 

Looking forward, our distinctive architecture paves the way for ubiquitous dietary monitoring. Future work will focus on deploying this framework on resource-constrained edge devices, such as smartphones and wearable glasses. This would enable real-time calorie tracking in the wild, significantly lowering the barrier to accurate personal health monitoring.